\documentclass{article}

\usepackage{arxiv}

\usepackage[utf8]{inputenc} 
\usepackage[T1]{fontenc}    
\usepackage[colorlinks=true,allcolors=blue]{hyperref}
\usepackage{url}            
\usepackage{booktabs}       
\usepackage{amsfonts}       
\usepackage{nicefrac}       
\usepackage{microtype}      
\usepackage{lipsum}		
\usepackage{graphicx}
\usepackage{natbib}
\usepackage{doi}
\usepackage{amsmath,subcaption}

\title{Reclaiming saliency: rhythmic precision-modulated action and perception}

\author{ {Ajith Anil Meera$^1$}\thanks{ Authors 1 and 2 have contributed equally to this work and share the first authorship.} \\
	Department of Cognitive Robotics, \\
	Faculty of Mechanical, Maritime and Materials \\
	Engineering,  Delft University of Technology, \\
    2628 CN Delft, The Netherlands\\
	\texttt{A.AnilMeera@tudelft.nl} \\
	\And
	{Filip Novicky$^2$} \\
	Donders Institute for Brain Cognition and Behavior,\\
	Department of Neurophysiology, \\
	Radboud University, The Netherlands \\
	\texttt{filip.novicky@donders.ru.nl} \\
	\And
	{Thomas Parr} \\
	Wellcome Centre for Human Neuroimaging, \\
	University College London, London, \\
	WC1N 3AR, United Kingdom\\
	\texttt{thomas.parr.12@ucl.ac.uk} \\
	\And
	{Karl Friston} \\
	Wellcome Centre for Human Neuroimaging, \\
	University College London, London, \\
	WC1N 3AR, United Kingdom\\
	\texttt{k.friston@ucl.ac.uk} \\
	\And
	{Pablo Lanillos} \\
	Donders Institute for Brain Cognition and Behavior,\\
	Department of Artificial Intelligence, \\
	Radboud University, The Netherlands \\
	\texttt{pablo.lanillos@donders.ru.nl} \\
	\And
	{Noor Sajid} \\
	Wellcome Centre for Human Neuroimaging, \\
	University College London, London, \\
	WC1N 3AR, United Kingdom \\
	\texttt{noor.sajid.18@ucl.ac.uk} \\
}

\renewcommand{\shorttitle}{Reclaiming saliency}

\hypersetup{
pdftitle={Reclaiming saliency: rhythmic precision-modulated action and perception},
pdfsubject={cs.RO, q-bio.NC},
pdfauthor={Ajith Anil Meera, Filip Novicky, Thomas Parr, Karl Friston, Pablo Lanillos and Noor Sajid},
pdfkeywords={Attention, Saliency, Free-energy principle, Active inference, Precision, Brain-inspired robotics, Cognitive robotics},
}

\begin{document}
\maketitle

\begin{abstract}
Computational models of visual attention in artificial intelligence and robotics have been inspired by the concept of a saliency map. These models account for the mutual information between the (current) visual information and its estimated causes. However, they fail to consider the circular causality between perception and action. In other words, they do not consider where to sample next, given current beliefs. Here, we reclaim salience as an active inference process that relies on two basic principles: uncertainty minimisation and rhythmic scheduling. For this, we make a distinction between attention and salience. Briefly, we associate attention with precision control, i.e., the
confidence with which beliefs can be updated given sampled sensory data, and salience with uncertainty minimisation that underwrites the selection of future sensory data. Using this, we propose a new account of attention based on rhythmic precision-modulation and discuss its potential in robotics, providing numerical experiments that showcase advantages of precision-modulation for state and noise estimation, system identification and action selection for informative path planning.
\end{abstract}


\section{Introduction}
Attention is a fundamental cognitive ability that determines which events from the environment, and the body, are preferentially  processed~\citep{itti2001computational}. For example, the motor system directs the visual sensory stream by orienting the fovea centralis (i.e., the retinal region of highest visual acuity)
towards points of interest within the visual scene. Thus, the confidence with which the causes of sampled
visual information are inferred is constrained by the physical structure of the eye – and eye movements
are necessary to minimise uncertainty about visual percepts 
~\citep{ahnelt1998photoreceptor}. In neuroscience, this can be attributed to two distinct, but highly interdependent attentional processes: ($i$) attentional gain mechanisms reliant on estimating the sensory precision of current data~\citep{feldman2010attention,yang2016active}, and ($ii$) attentional salience that involves actively engaging with the sensorium to sample appropriate future data~\citep{parr2019attention,lengyel2016active}. Put simply, we formalise the fundamental difference between attention –- as optimising perceptual processing -– and salience as optimising the sampling of what is processed. This highlights the dynamic, circular nature with which biological agents acquire, and process, sensory information.

Understanding the computational mechanisms that undergird these two attentional phenomena is pertinent for deploying apt models of (visual) perception in artificial agents~\citep{klink2014priority,mousavi2016learning,atrey2019exploratory} and robots~\citep{frintrop2008attentional,begum2010visual,ferreira2014attentional,lanillos2015designing}. Previous computational models of visual attention, used in artificial intelligence and robotics, have been inspired (and limited) by the feature integration theory proposed by~\citep{treisman1980feature} and the concept of a saliency map~\citep{tsotsos1995modeling,itti2001computational,borji2012state}. Briefly, a saliency map is a static two-dimensional `image' that encodes stimulus relevance, e.g., the importance of particular region. These maps are then used to isolate relevant information for control (e.g., to direct foveation of the maximum valued region). Accordingly, computational models reliant on this formulation do not consider the circular-dependence between action selection and cue relevance -- and simply use these static saliency maps to guide action. 

In this article, we adopt a first principles account to disambiguate the computational mechanisms that underpin attention and salience~\citep{parr2019attention} and provide a new account of attention. Specifically, our formulation can be effectively implemented for robotic systems and facilitates both state-estimation and action selection. For this, we associate attention with precision control, i.e., the confidence with which beliefs can be updated given (current) sampled sensory data. Salience is associated with uncertainty minimisation that influences the selection of future sensory data. This formulation speaks to a computational distinction between action selection (i.e., where to look next) and visual sampling (i.e., what information is being processed). Importantly, recent evidence demonstrates the rhythmic nature of these processes via a theta-cycle coupling that fluctuates between high and low precision---as unpacked in Sec~\ref{sec:neuroscience}. From a robotics perspective, resolving uncertainty about states of affair speaks to a form of Bayesian optimality, in which decisions are made to maximise expected information gain~\citep{lindley1956measure,friston2021sophisticated,sajid2021efe}. The duality between attention and salience is important for resolving uncertainty and enabling active perception. Significantly, it addresses an important challenge for defining autonomous robotics systems that can balance optimally between data assimilation (i.e., confidently perceiving current observations) and exploratory behaviour to maximise information gain~\citep{bajcsy2018revisiting}. 

In what follows, we review the neuroscience of attention and salience (Sec. \ref{sec:neuroscience}) to develop a novel (computational) account of attention based on precision-modulation that underwrites perception and action (Sec. \ref{sec:salience-precision}). Next, we face-validate our formulation within a robotics context using numerical experiments (Sec. \ref{sec:robotics}). The robotics implementation instantiates a free energy principle (FEP) approach to information processing~\citep{friston2010free}. This allows us to modulate the (appropriate) precision parameters to solve relevant robotics challenges in perception and control; namely, state-estimation (Sec.~\ref{sec:noise_filtering}), system identification (Sec.~\ref{sec:system_ID}), planning (Sec.~\ref{sec:rob:actions}), and active perception (Sec.~\ref{sec:act_per_coupling}). We conclude with a discussion of the requisite steps for instantiating a full-fledged computational model of precision-modulated attention – and its implications in a robotics setting.


\section{Attention and salience in neuroscience} \label{sec:neuroscience}

Our interactions with the world are guided by efficient gathering and processing of sensory information. The quality of these acquired sensory data is reflected in attentional resources that select sensations which influence our beliefs about the (current and future) states of affairs~\citep{lengyel2016active,yang2016theoretical}. This selection is often related to gain control, i.e., an increase of neural spikes when an object is attended to. However, gain control only accounts for half the story because we can only attend to those objects that are within our visual field. Accordingly, if a salient object is outside the centre of our visual field, we orient the fovea to points of interest. This involves two separate, but often conflated, processes: attention and salience -- where the former relates to processing current visual data, and the latter to ensuring the agent samples salient data in the future~\citep{parr2019attention}. That these two processes are strongly coupled is exemplified by the pre-motor theory of attention ~\citep{rizzolatti1987reorienting}, which highlights the close relationship between overt saccadic sampling of the visual field and the covert deployment of attention in the absence of eye movements. Specifically, it posits that covert attention\footnote{Covert attention is where saccadic eye movements do not occur.} is realised via processes that are generated by particular eye movements but inhibits the action itself. In this sense, it does not distinguish between covert and
overt\footnote{Overt attention deals with how an agent tracks the object with eye movements} types of attention. 

From a first principles (Bayesian) account, it is necessary to separate between attention and salience because they speak to different optimisation processes. Explicitly, attention as a precision-dependent (neural) gain control mechanism that facilitates optimisation of the \textit{current} sampled sensory data~\citep{feldman2010attention,desimone1996neural}. Conversely, salience is associated with selection of \textit{future} data that reduces uncertainty~\citep{berk2016,karlepistemic,parr2019attention}. Put simply, it is possible to optimise attention in the absence of eye movements and active vision, whereas salience is necessary to optimise the deployment of eye movements. In what follows, we formalise this distinction with a particular focus on visual attention~\citep{kanwisher2000visual}, and discuss recent findings that speak to a rhythmic coupling that underwrites periodic deployment of gain control and saccades, via modulation of distinct precision parameters. 

\subsection{Attention as neural gain control}\label{sec::neuro_precision}
Neural gain control (or precision) can be regarded as an amplifier of neural communication during attention tasks~\citep{eldar2013effects,reynolds2000attention}. Specifically, an increase in gain amplifies the postsynaptic responses of neurons to their pre-synaptic input. Thus, gain control rests on synaptic modulation that can emphasise -– or preferentially select -– a particular type of sensory data. From a Bayesian perspective
~\citep{rao2005bayesian,spratling2008predictive,parr2018precision}, this speaks to the confidence with which beliefs can be updated given sampled sensory data (i.e., optimal state estimation) -- under a generative model~\citep{whiteley2008implicit,parr2018precision}. For example, affording high precision to certain sensory inputs would lead to confident Bayesian belief updating. However, low precision reduces the influence of sensory input by attenuating the precision of the likelihood, relative to a prior belief, and current observations would do little to resolve ensuing uncertainty. Thus, sampled visual data (from different areas) can be predicted with varying levels of precision, where attention accentuates sensory precision. The deployment of precision or attention is influenced by competition between stimuli (i.e., which sensory data to sample) and prior beliefs. Interestingly, casting attention as precision or, equivalently, synaptic gain offers a coherency between biased competition~\citep{desimone1996neural}, predictive coding~\citep{spratling2008predictive} and generic active inference schemes~\citep{feldman2010attention,parr2018precision,brown2013active,kanai2015cerebral}. 

Naturally, gain control is accompanied by neuronal variability, i.e., sharpened neural responses for the same task over time. Consistent with gain control, these fluctuations in neural responses across trials can be explained by precision engineered message passing ~\citep{clark2013precision} via ($i$) normalization models~\citep{reynolds2009normalization,ruff2016stimulus}, ($ii$) temperature parameter manipulation~\citep{mirza2019introducing,parr2018precision,parr2017uncertainty,parr2019perceptual,feldman2010attention}, or ($iii$) introduction of (conjugate hyper-)priors that are either pre-specified~\citep{sajid2021bayesian,sajid2020neuromodulatory} or optimised using uninformed priors~\citep{friston2003dynamic,anil2021dynamic}. Recently, these approaches have been used to simulate attention by accentuating predictions about a given visual stimulus \citep{reynolds2009normalization,feldman2010attention,ruff2016stimulus}. For example, normalization models propose that every neuronal response is normalized within its neuronal ensemble (i.e., the surrounding neuronal responses)~\citep{heeger1992normalization,louie2019normalization}. Thus, to amplify the neuronal response of particular neuron, the neuronal pool has to be inhibited such that that particular neuron has a sharper evoked response~\citep{schmitz2018normalization}. Importantly, these (superficially distinct) formulations simulate similar functions using different procedures to accentuate responses over a particular neuronal pool for a given neuron or a group of neurons. This introduces shifts in precision to produce attentional gain and the precision of neuronal encoding.

\subsection{Salience as uncertainty minimisation}
In the neurosciences, (visual) salience refers to the `significance' of particular objects in the environment. Salience often implicates the superior colliculus, a region that encodes eye movements~\citep{white2017superior}. This makes intuitive sense, as the superior colliculus plays a role in generation of eye movements -- being an integral part of the brainstem oculomotor network~\citep{raybourn1977colliculoreticular} -- and salient objects provide information that is best resolved in the centre of the visual field, thus motivating eye movements to that location. For this reason, our understanding of salience is a quintessentially action-driving phenomenon~\citep{parr2019attention}. Mathematically,
salience has been defined as Bayesian surprise~\citep{itti2001computational,itti2009bayesian}, intrinsic motivation~\citep{oudeyer2009intrinsic}, and subsequently, epistemic value under active inference~\citep{mirza2016scene,parr2018precision}. Active inference -- a Bayesian account of perception and action~\citep{friston2017active,da2020active} -- stipulates that action selection is determined by uncertainty minimisation. Formally, uncertainty minimisation speaks to minimisation of an expected free energy functional over future trajectories~\citep{da2020active,sajid2021efe}. This action selection objective can be decomposed into epistemic and extrinsic value, where the former pertains to exploratory drives that encourage resolution of uncertainty by sampling salient observations, e.g., only checking one's watch when one does not know the time. However, after checking the watch there is little epistemic value in looking at it again. Generally, the tendency to seek out new locations -- once uncertainty has been resolved at the current fixation point -- is called inhibition of return~\citep{klein2000inhibition}. 

From an active inference perspective, this phenomenon is prevalent because a recent action has already resolved the uncertainty about the time and checking again would offer nothing more in terms of information gain~\citep{parr2019attention}. Accordingly, salience involves seeking sensory data that have a predictable, uncertainty reducing, effect on current beliefs about states of affairs in the world~\citep{parr2018precision,mirza2016scene}. Thus salience contends with beliefs about data that must be acquired and the precision of beliefs about policies (i.e., action trajectories) that dictate it. Formally, this emerges from the imperative to maximise the amount of information gained regarding beliefs, from observing the environment. Happily, prior studies have made the connection between eye movements, salience, and precision manipulation~\citep{friston2011action,brown2013active,crevecoeur2017saccadic}. This connection emerges from planning strategies that allow the agent to minimise uncertainty by garnering the right kind of data. 

Next, we consider recent findings on how the coupling of these two mechanisms, attention and salience, may be realised in the brain. 

\subsection{Rhythmic coupling of attention and salience}\label{sec::neuron_coupling}

To illustrate the coupling between attention and salience, we turn to a recent rhythmic theory of attention. The theory proposes that coupling of saccades, during sampling of visual information, happens at neuronal and behavioural theta oscillations;  a frequency of 3-8Hz~\citep{fiebelkorn2019rhythmic,fiebelkorn2021spike}. This frequency simultaneously allows for: ($i$) a systematic integration of visual samples with action, and ($ii$) a temporal schedule to disengage and search the environment for more relevant information. 

Given that gain control is related to increased sensory precision, we can accordingly relate saccadic eye movements to the decreased precision. This introduces saccadic suppression, a phenomenon that decreases visual gain during eye movements \citep{crevecoeur2017saccadic}. This phenomenon was described by
Helmholtz who observed that externally initiated eye movements (e.g., when oneself gently presses a side of an eye) eludes the saccadic suppression that accompanies normal eye movements – and we see the world shift, because optic flow is not attenuated~\citep{helmholtz1925handbook}. An interesting consequence of this is that, as eye movements happen periodically~\citep{rucci2018temporal,benedetto2020common}, there must be a periodic switch between high and low sensory precision, with high precision (or enhanced gain) during fixations and low precision (or suppressed gain) during saccades. Interestingly, it has been shown that rather than having action resetting the neural periodicity, it is better understood as something that aligns within an already existing rhythm \citep{tomassini2017theta,hogendoorn2016voluntary}. Additionally, the rhythmicity of higher and lower fidelity of sensory sampling has been
shown to fluctuate rhythmically around 3Hz ~\citep{benedetto2017saccadic}, suggesting that action emerges rhythmically when visual precision is low~\citep{hogendoorn2016voluntary}, triggering salience.

Building upon this, we hypothesise that theta rhythms generated in the fronto-parietal network \citep{fiebelkorn2018dynamic,helfrich2018neural,fiebelkorn2020functional} couples saccades with saccadic suppression causing the switches between visual sampling and saccadic shifting. This introduces a diachronic aspect to the belief updating process~\citep{FRISTON202042,parr2021understanding,Sajid2022diachronic}; i.e., sequential fluctuations between attending to current data (perception) and seeking new data (action). This supports empirical findings that both eye movements~\citep{sommer2006influence} and filtering irrelevant information~\citep{nakajima2019prefrontal,phillips2016subcortical,fiebelkorn2020functional} are initiated in this cortical network. Interestingly, both eye movements and visual filtering then propagate to sub-cortical regions, i.e., the superior colliculus---for saliency map composition \citep{white2017superior}---and the thalamus---for gain control \citep{kanai2015cerebral,fiebelkorn2019mediodorsal}, respectively. Furthermore, this is consistent with recent findings that the periodicity of neural responses are important for understanding the relation of motor responses and sensory information -- i.e., perception-action coupling~\citep{benedetto2020common}. Importantly, theta rhythms also speak to the speed (i.e., the temporal schedule) with which visual information is sampled from the environment~\citep{busch2010spontaneous,dugue2015theta,dugue2016attention,helfrich2018neural}. Meaning visual information is not sampled continuously, as our visual experiences would suggest, but rather it is made of successive discrete samples~\citep{vanrullen2016perceptual,parr2021generative}. 

In summary, the computations that underwrite attention and active vision are coupled and exhibit circular causality. Briefly, selective attention and sensory attenuation optimize the processing of sensory samples and which particular visual percepts are inferred. In turn, this determines appropriateness of future eye movements (or actions). Interestingly, the close functional (and
computational) link between the two mechanisms endorses the pre-motor theory of attention.

\section{Proposed precision-modulated account of attention and salience}
\label{sec:salience-precision}

\begin{figure}[!t]
    \hfill
    \begin{center}
    \includegraphics[scale=0.6]{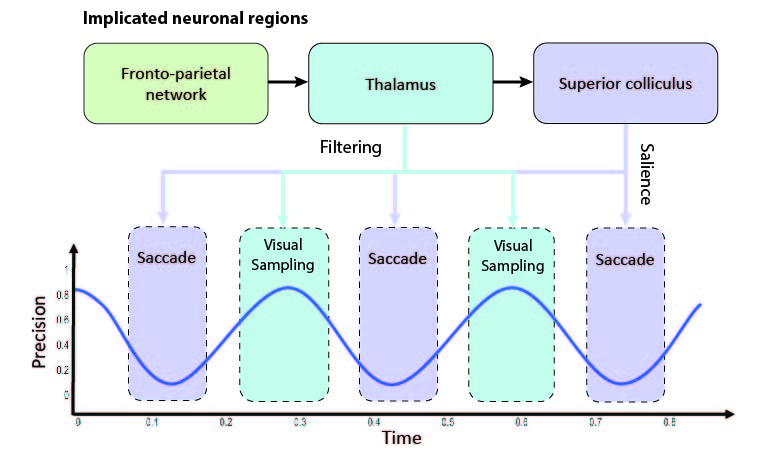}
    \end{center}
    \vspace{-8mm}
    \caption{A graphical illustration of the precision-modulated account of perception and action. Salience and attention are computed based upon beliefs (assumed to be) encoded in parts of the fronto-parietal network and realised in distinct brain regions: superior colliculus (SC) for perception as inference and thalamus for planning as inference, respectively. To deploy attentional processes efficiently, these two mechanisms have to be aligned, which is done rhythmically, hypothetically in theta frequency. This coupling enables the saccadic suppression phenomenon through fluctuations in precision (on an arbitrary scale). When precision is low (i.e., the trough of the theta rhythm), the saccade emerges. Note that there might be distinct processes inhibiting the action (e.g., covert attention). On the other hand, high precision facilitates confident inferences about the causes of visual data. Under this account, thalamus is used for initiating gain control (or visual sampling in general) by providing stronger sensory input, while superior colliculus dictates next saccades, that lead to most informative fixation positions.}
      \label{fig:summary_fig}
\end{figure}

Here, we introduce our precision-modulated account of perception and action. A graphical illustration is provided in  Figure~\ref{fig:summary_fig}. For this, we turn to attention and salient action selection which have their roots in biological processes relevant for acquiring task-relevant information. Under an active inference account, this attention influences (posterior) state estimation and can be associated with increased precision of belief updating and gain control---described in Sec.~\ref{sec::neuro_precision}. Furthermore, this is distinct from salience despite interdependent neuronal composition and computations. 

Further alignment between the two constructs can be revealed by considering the temporal scheduling between movement (i.e., action) and perception for uncertainty resolution~\citep{parr2019attention}. 
We postulate that this perception-action coupling is best understood as a periodic fluctuation between minimising uncertainty and precision control. Subsequently, action is deployed to reduce uncertainty.
Such an alignment specifies what stimulus is selected and under what level of precision it is processed. \cite{parr2019attention} hypothesise that action alignment with precision is due to the eye structure that provides precise information in the fovea and requires the agent to foveate the most informative stimulus. We extend this by considering the periodic deployment of gain control with saccades~\citep{hogendoorn2016voluntary,fiebelkorn2019rhythmic,nakayama2019attention,tomassini2017theta,benedetto2017saccadic}. 

Accordingly, our formulation defines attention as precision control and salience as uncertainty minimisation supported by discrete sampling of visual information at a theta rhythm. This synchronises  perception and action together in an oscillatory fashion~\citep{hogendoorn2016voluntary}. Importantly, a Bayesian formulation of this can be realised as precision manipulation over particular model parameters. We reserve further details for Sec.~\ref{sec:robotics}.

\noindent\textbf{Summary} Based upon our review, we propose a precision-modulated account of attention and salience, emphasising the diachronic realisation of action and perception. In the following sections, we investigate a realisation of this model for a robotic system.

\section{Precision-based attention for Robotics}
\label{sec:robotics}

The previous section introduced a conceptual account to explain the computational mechanisms that undergird attention based on neuroscience findings. We focused on reclaiming saliency as an active process that relies on neural gain control, uncertainty minimisation and structured scheduling. Here, we describe how we can mathematically realise some of these mechanisms in the context of well-known challenges in robotics. Enabling robots with this type of attention may be crucial to filter the sensory signals and internal variables that are relevant to estimate the robot/world state and complete any task. More importantly, the active component of salience (i.e., behaviour) is essential to interact with the world---as argued in active perception approaches~\citep{bajcsy2018revisiting}. 

We revisit the standard view of attention in robotics by introducing sensory precision (inverse variance) as the driving mechanism for modulating both perception and action~\citep{clark2013precision,friston2011action}. Although saliency was originally described to underwrite behaviour, most models used in
robotics, strongly biased by computer vision approaches, focus on computing the most relevant region
of an image ~\citep{borji2012state}---mainly computing human fixation maps---relegating action to a secondary process. Illustratively, state-of-the-art deep learning saliency models---as shown in the MIT saliency benchmark~\citep{mit-saliency-benchmark}---do not have the action as an output. Conversely, the active perception approach properly defines the action as an essential process of active sensing to gather the relevant information. Our proposed model, based on precision modulated action and perception coupling ($i$) place attention as essential for state-estimation and system identification and ($ii$) and reclaims saliency as a driver for information-seeking behaviour, as proposed in early works~\citep{tsotsos1995modeling}, but goes beyond human fixation maps for both improving the model of the environment (exploration) and solving the task (exploitation).


\begin{table}[!htb]
	\caption{Robotics applications and their precision realisations.}
	\centering
	\begin{tabular}{llll}
		\toprule
		Task & Application & Precision manipulation & Section \\
		\midrule
		Perception & State \& input estimation & Noise precision modelling $\tilde{\Pi}$ & \ref{sec:noise_filtering} \\
          & System Identification & Posterior parameter precision learning $\Pi^\theta$ & \ref{sec:system_ID} \\
          & Exploration-exploitation in learning & Prior parameter precision modelling $P^\theta$ & \ref{sec:exploration_expl} \\ 
          & Noise estimation & Noise precision learning $\tilde{\Pi}$ & \ref{sec:noise_prec_learn} \\
          \midrule
          Action & Informative Path Planning (IPP) & Precision optimisation (of map) &  \ref{sec:rob:actions:optimization} \\
          \midrule
          Active perception & IPP with   action-perception cycle  & Precision modulation & \ref{sec:act_per_coupling} \\
		\bottomrule
	\end{tabular}
	\label{tab:prec_modulations}
\end{table}

In what follows, we highlight the key role of precision by reviewing relevant brain-inspired attention models deployed in robotics (Sec.~\ref{sec:saliencyrobotics}). We propose precision-modulated attentional  mechanisms for robots in three contexts - perception (Sec.~\ref{sec:rob:model}), action (Sec.~\ref{sec:rob:actions}) and active perception (Sec.~\ref{sec:act_per_coupling}). 
The precision-modulated perception is formalised for a robotics setting; via ($i$) state estimation (i.e., estimating the hidden states of a dynamic system from sensory signals -- Sec.~\ref{sec:noise_filtering}), and ($ii$) system identification (i.e., estimating the parameters of the dynamic system from sensory signals -- Sec.~\ref{sec:system_ID}). Next, we show that precision-modulated action can be realised through precision optimisation (planning future actions -- Sec.~\ref{sec:rob:actions:optimization}) and discuss practical considerations for coupling with precision-modulated perception (precision based active perception - Sec.~\ref{sec:act_per_coupling}). Table~\ref{tab:prec_modulations} summarises our proposed precision manipulations to solve relevant problems in robot perception and action. Table~\ref{tab:prec_definitions} provides the definitions of precision within our mechanism. 


\begin{table}[!htb]
	\caption{Precision parameters that are manipulated in ~Sec.\ref{sec:rob:model}}
	\centering
	\begin{tabular}{lll}
		\toprule
		Term & Symbol & Definition  \\
		\midrule
		Sensory precision & $\Pi^z$ & Inverse covariance of sensory noise $\textbf{z}$ (Eqn.~\ref{eqn:general_LTI}). \\
         Prior parameter precision & $P^\theta$ & The robot's confidence on its prior parameters $\eta^\theta$. \\
         Noise precision & $\tilde{\Pi}$  & The inverse covariance of all noises (Eqn.~\ref{eqn:precision_generalized}).  \\
         Posterior parameter precision & $\Pi^\theta$ & The robot's confidence on its parameter estimates. \\
		\bottomrule
	\end{tabular}
	\label{tab:prec_definitions}
\end{table}

\subsection{Previous brain-inspired attention models in robotics}
\label{sec:saliencyrobotics}
Brain-inspired attention has been mainly addressed in robotics from a `passive' visual saliency perspective, e.g., which pixels of the image are the most relevant. This saliency map is then generally used to foveate the most salient region. This approach was strongly influenced by early computational models of visual attention~\citep{tsotsos1995modeling,itti2001computational}. The first models deployed in robots were bottom-up, where the sensory input was transformed into an array of values that represents the importance (or salience) of each cue. Thus, the robot was able to identify which region of the scene has to look at, independently of the task performed---see \cite{borji2012state} for a review on visual saliency. These models have also been useful for acquiring meaningful visual features in applications, such as object recognition~\citep{orabona2005object,frintrop2006vocus}, localisation, mapping and navigation~\citep{frintrop2008attentional,kim2013real,roberts2012saliency}. Saliency computation was usually employed as a helper for the selection of the relevant characteristics of the environment to be encoded. Thus, reducing the information needed to process.

More refined methods of visual attention employed top-down modulation, where the context, task or goal bias the relevance of the visual input. These methods were used, for instance, to identify humans using motion patterns~\citep{butko2008visual,moren2008biologically}. A few works also focused on object/target search applications, where top-down and bottom-up saliency attention were used to find objects or people in a search and rescue scenario~\citep{rasouli2020attention}. 

Attention has also been considered in human-robot interaction and social robotics applications~\citep{ferreira2014attentional}, mainly for scene or task understanding~\citep{ude2005distributed,kragic2005vision,lanillos2016yielding}, and gaze estimation~\citep{shon2005probabilistic} and generation~\citep{lanillos2015designing}. For instance, computing where the human is looking at and where the robot should look at or which object should be grasped. Furthermore, multi-sensory and 3D saliency computation has also been investigated~\citep{lanillos2015multisensory}. Finally, more complex attention behaviours, particularly designed for social robotics and based on human non-verbal communication, such as joint attention, have also been addressed. Here the robot and the human share the attention of one object through meaningful saccades, i.e., head/eye movements~\citep{kaplan2006challenges,nagai2003constructive,lanillos2015designing}.

Although attention mechanisms have been widely investigated in robotics, specially to model visual cognition~\citep{begum2010visual,kragic2005vision}, the majority of the works have treated attention as an extra feature that can help the visual processing, instead of a crucial component needed for the proper functioning of the cognitive abilities of the robot~\citep{lanillos2018activeattention}. Furthermore, these methods had the tendency to leave the action generation out of the attention process. One of the reasons for  not including saliency computation, in robotic systems, is that the majority of the models only output ‘human-fixation map’ predictions, given a static image. Saliency computation introduces extra computational complexity, which can be finessed by visual segmentation algorithms (e.g., line detectors in autonomous navigation). However, it does not resolve uncertainty nor select actions that maximise information gain in the future. In essence, the incomplete view of attention models that output human-fixation maps has arguably obscured the huge potential of neuroscience-inspired attentional mechanisms for robotics.

Our proposed model of attention, based on precision modulation, abandons the current robotics narrow view of attention and saliency by explicitly modelling attention within state estimation, learning and control. Thus, placing attentional processes at the core of the robot computation and not as an extra add-on. In the following sections, we describe the realisation of our precision-based attention formulation in robotics using common practical applications as the backbone motif.

\subsection{Precision-modulated perception}
\label{sec:rob:model}
We formalise precision-modulated perception from a first principles Bayesian perspective -- explicitly the free energy principle approach proposed by~\cite{friston2011action}. Practically, this entails optimising precision parameters over (particular) model parameters. 

Through numerical examples we show how our model is able to perform accurate state estimation~\citep{bos2021free} and stable parameter learning~\citep{meera2021brain,10.1007/978-3-030-93736-2_49}. To illustrate the approach, we first introduce  a dynamic system modelled as a linear state space system in robotics (Sec.~\ref{sec:prec_state_space})---we used this formulation in all our numerical experiments. We briefly review the formal terminologies for a robotics context to appropriately situate our precision-based mechanism for perception. Explicitly, we introduce: precision modelling (by adapting a known form of the precision matrix), precision learning (by learning the full precision matrix), and precision optimisation (use precision as an objective function during learning). As a reminder, precision modelling is associated with (instantaneous) gain control and precision learning (at slower time scales) is associated with optimising that control.

\subsubsection{Precision for state space models} 
\label{sec:prec_state_space}
A linear dynamic system can be modelled using the following state space equations (boldface notation denotes components of the real system and non-boldface notation its estimates):
\begin{equation} 
\label{eqn:general_LTI}
    \begin{split}
       \Dot{\textbf{x}} = \textbf{A} \textbf{x}+ \textbf{B} \textbf{u} + \textbf{w},
    \end{split}
    \quad \quad
    \begin{split}
        \textbf{y} = \textbf{C} \textbf{x} +\textbf{z}.
    \end{split}{}
\end{equation}
where $\textbf{A}$, $\textbf{B}$ and $\textbf{C}$ are constant matrices defining the system parameters, $\textbf{x}\in \mathbb{R}^n$ is the system state (usually an unobserved variable), $\textbf{u}\in \mathbb{R}^r$ is the input or control actions, $\textbf{y}\in \mathbb{R}^m$ is the output or the sensory measurements, $\textbf{w}\in \mathbb{R}^n$ is the process noise with precision $\mathbf{\Pi^w}$ (or inverse variance $\mathbf{{\Sigma^w}}^{-1}$), and $\textbf{z}\in \mathbb{R}^m$ is the measurement noise with precision $\mathbf{\Pi^z}$.


For instance, we can describe a mass-spring damper system (depicted in Fig.~\ref{fig:generalized_coordinates}b) using state space equations. A mass ($m=1.4 kg$) is attached to a spring with elasticity constant ($k=0.8 N/m$), and a damper with a damping coefficient ($b = 0.4 Ns/m$). When a force ($u(t) = e^{-0.25(t-12)^2}$) is applied on the mass, it displaces $x$ from its equilibrium point. The linear dynamics of this system is given by:
\begin{equation} \label{eqn:spring_damper_LTI}
\begin{split}
    \begin{bmatrix}
    \dot{x} \\ \ddot{x}
    \end{bmatrix} & = \begin{bmatrix}
    0 & 1 \\ -\frac{k}{m} & -\frac{b}{m}
    \end{bmatrix} \begin{bmatrix}
    x \\ \dot{x}
    \end{bmatrix} + \begin{bmatrix}
    0 \\ \frac{1}{m}
    \end{bmatrix} u,\end{split}
    \quad \quad
    \begin{split}
        y & = \begin{bmatrix}
        1 & 0
        \end{bmatrix} \begin{bmatrix}
    x \\ \dot{x}
    \end{bmatrix}.
    \end{split}
\end{equation}
Note that Eq. (\ref{eqn:spring_damper_LTI}) is equivalent to Eq. (\ref{eqn:general_LTI}) with parameters \textbf{A} $ = \begin{bmatrix}
    0 & 1 \\ -\frac{k}{m} & -\frac{b}{m}
    \end{bmatrix},$ \textbf{B} =$ \begin{bmatrix}
    0,  \frac{1}{m}
    \end{bmatrix}^T$ and \textbf{C} $=  \begin{bmatrix}
        1 & 0
        \end{bmatrix}, $ and state {x} = $\begin{bmatrix}
    x, \dot{x}
    \end{bmatrix}^T$.

Now we introduce attention as precision modulation assuming that the robotic goal is to minimise the prediction error~\citep{friston2011action,lanillos2018adaptive,meera2020free}, i.e., to refine its model of the environment and perform accurate state estimation, given the information available. In other words, the robot has to estimate \textbf{x} and \textbf{u} from input prior $\eta^u$ with a prior precision of $P^u$, given the measurements \textbf{y}, parameters \textbf{A}, \textbf{B}, \textbf{C} and noise precision $\mathbf{\Pi^w}$ and $\mathbf{\Pi^z}$. Formally, the prediction error $\tilde{\epsilon}$ of the sensory measurements $\tilde{\epsilon}^y $, control input reference $\tilde{\epsilon}^u $ and state $\tilde{\epsilon}^x$ are:
\begin{equation} \label{eqn:epsilon_precision}
\tilde{\epsilon} = \begin{bmatrix}
\tilde{\epsilon}^y \\ \tilde{\epsilon}^u \\ \tilde{\epsilon}^x
\end{bmatrix} = \begin{bmatrix}
\tilde{\textbf{y}} - \tilde{\textbf{C}}\tilde{x} \\ \tilde{u} - \tilde{\eta}^u \\ D^x \tilde{x} - \tilde{\textbf{A}}\tilde{x} - \tilde{\textbf{B}} \tilde{u}
\end{bmatrix}
\quad
\begin{cases}
     \text{sensory prediction error}\\
    \text{control input prediction error}\\
    \text{state prediction error}
\end{cases}
\end{equation}
Note that $\tilde{\epsilon}^y = \tilde{\textbf{y}} - \tilde{\textbf{C}}\tilde{x}$ is the difference between the observed measurement and the predicted sensory input given the state\footnote{The tilde over the variable refers to the generalised coordinates, i.e., the variable includes all temporal derivatives. Thus, $\tilde{\epsilon}$ is the combined prediction error of outputs, inputs and states. For example, the generalised output $\tilde{y}$ is given by $\tilde{\textbf{y}} = [y, y' , y''...]^T$, where the prime operator denotes the derivatives. We use generalised coordinates~\citep{friston2010generalised} for achieving accurate state and input estimation during the presence of (coloured) noise by modelling the time dependent quantities ($x,v,y,w,z$) in generalised coordinates. This involves keeping track of the evolution of the trajectory of the probability distributions of states, instead of just their point estimates. Here the coloured noise $w$ and $z$ are modelled as a white noise convoluted with a Gaussian kernel. The use of generalised coordinates has recently shown to outperform classical approaches under coloured noise on real quadrotor flight \citep{bos2021free}}. Here $D^x$ performs the (block) derivative operation, which is equivalent to shifting up all the components in generalised coordinates by one block. 

We can estimate the state and input using the Dynamic Expectation Maximisation (DEM) algorithm~\citep{friston2008variational,meera2020free} that optimises a free energy variational bound $\mathcal{F}$ to be tractable\footnote{Note that this expression of the variational free energy is using the Laplace and mean-field approximations commonly used in the FEP literature}. This is:

\begin{equation} \label{eqn:F_state}
   X = \begin{bmatrix}
   \tilde{x} \\ \tilde{u}
   \end{bmatrix} =  \arg\max_{X}  \mathcal{F} = \arg\max_{X}  -\frac{1}{2} \tilde{\epsilon}^T \tilde{\Pi} \tilde{\epsilon}
\end{equation}
Crucially, $\tilde{\Pi}$ is the generalised noise precision that modulates the contribution of each prediction error to the estimation of the state and the computation of the action. Thus, $\tilde{\Pi}$ is equivalent to attentional gain. For instance, we can model the precision matrix to attend to the most informative signal derivatives in $\tilde{\textbf{y}}$. Concisely, the precision $\tilde{\Pi}$ has the following form:
\begin{equation} \label{eqn:precision_generalized}
    \tilde{\Pi} = \begin{bmatrix}
    S \otimes \Pi^z & 0 & 0 \\ 0 & S \otimes P^u & 0 \\ 0 & 0 & S \otimes \Pi^w
    \end{bmatrix},
\end{equation}
where $S$ is the smoothness matrix. In Sec. \ref{sec:noise_filtering}, we show that modelling the precision matrix $\tilde{\Pi}$ using the $S$ matrix improves the estimation quality. 

The full free energy functional (time integral of free energy $\Bar{\mathcal{F}} = \int \mathcal{F} dt$ at optimal precision) that the robot optimises to perform state-estimation and system identification is described in Eq. (\ref{eqn:F_action_opt})---for readability we omitted the details of the derivation of this cost function, and we refer to \citep{anil2021dynamic} for further details.

\begin{equation} 
\label{eqn:F_action_opt}
    \begin{split}
        \Bar{\mathcal{F}}  = &    - \frac{1}{2} \sum_t \Big[ \underbrace{ \tilde{\epsilon}^{yT} \tilde{\Pi}^z  \tilde{\epsilon}^y + \tilde{\epsilon}^{uT} P^{\tilde{u}} \tilde{\epsilon}^u  + \tilde{\epsilon}^{xT} \tilde{\Pi}^w \tilde{\epsilon}^x }_{\text{precision weighed prediction error}} \Big]  - \frac{1}{2} \Big[ \underbrace{ {\epsilon}^{\theta T} P^\theta {\epsilon}^\theta + {\epsilon}^{\lambda T} P^\lambda {\epsilon}^\lambda}_{\text{prior precision weighed prediction error of $\theta$ and $\lambda$}} \Big] \\
                &     + \underbrace{\frac{1}{2} n_t \ln |\Sigma^X|}_{\text{state and input entropy}}     + \frac{1}{2} n^t \underbrace{ \big[  \ln |\tilde{\Pi}^z| +  \ln |P^{\tilde{v}}| +  \ln |\tilde{\Pi}^w| \big] }_{\text{noise entropy}}  + \underbrace{\frac{1}{2} \ln |\Sigma^\theta P^\theta|}_{\text{parameter entropy}}
                    +\underbrace{\frac{1}{2} \ln |\Sigma^\lambda P^\lambda|}_{\text{hyperparameter entropy}}  \\ 
    \end{split}
\end{equation}
Here $\epsilon^\theta = \theta - \eta^\theta$, $\epsilon^\lambda = \lambda - \eta^\lambda$ are the prediction errors of parameters and hyper-parameters\footnote{System identification involves the estimation of system parameters (denoted by $\theta$, e.g., vectorised $\textbf{A}$), given $\textbf{y}, \textbf{u}$, by starting from a parameter prior of $\eta^\theta$ with prior precision $P^\theta$, and a prior on noise hyper-parameter $\eta^\lambda$ with a prior precision of $P^\lambda$. Note that we parametrise noise precision ($\Pi^w$ and $\Pi^z$) using $\lambda \in \mathbb{R}^{2 \times 1} = \big[ \begin{smallmatrix} \lambda^z \\ \lambda^w \end{smallmatrix} \big]$ as an exponential relation  (e.g., $\Pi^w(\lambda^w) = \exp(\lambda^w) I^{n \times n}$).}. $\Bar{\mathcal{F}}$ consist of two main components: i) precision weighed prediction errors and ii) precision-based entropy. The dominant role of precision –- in the free energy objective -– is reflected in how modulating these precision parameters can have a profound influence on perception and behaviour. The theoretical guarantees for stable estimation \citep{10.1007/978-3-030-93736-2_49}, and its application on real robots \citep{lanillos2021-aif-challenges} make this formulation very appealing to robotic systems. 

Note that we can manipulate three kinds of precision within the state space formulation: i) prior precision ($P^{\tilde{u}},P^\theta,P^\lambda$), ii) conditional precision on estimates ($\Pi^{X},\Pi^\theta,\Pi^\lambda$) and iii) noise precision ($\Pi^z,\Pi^w$). Therefore, to learn the correct parameter values $\theta$, we i) learn the parameter precision $\Pi^\theta$, ii) model the prior parameter precision $P^\theta$, and iii) learn the noise precision $\Pi^w$ and $\Pi^z$ (parameterised using $\lambda$).

\subsubsection{State and input estimation} \label{sec:noise_filtering}
State estimation is the process of estimating the unobserved states of a real system from (noisy) measurements. Here, we show how we can achieve accurate estimation through precision modulation in a linear time invariant system under the influence of coloured noise \citep{meera2020free}. State estimation in the presence of coloured noise is inherently challenging, owing to the non-white nature of the noise, which is often ignored in conventional approaches, such as the Kalman Filter~\citep{welch1995introduction}.

\begin{figure}[!htb]
\centering 
\includegraphics[width=0.8\textwidth]{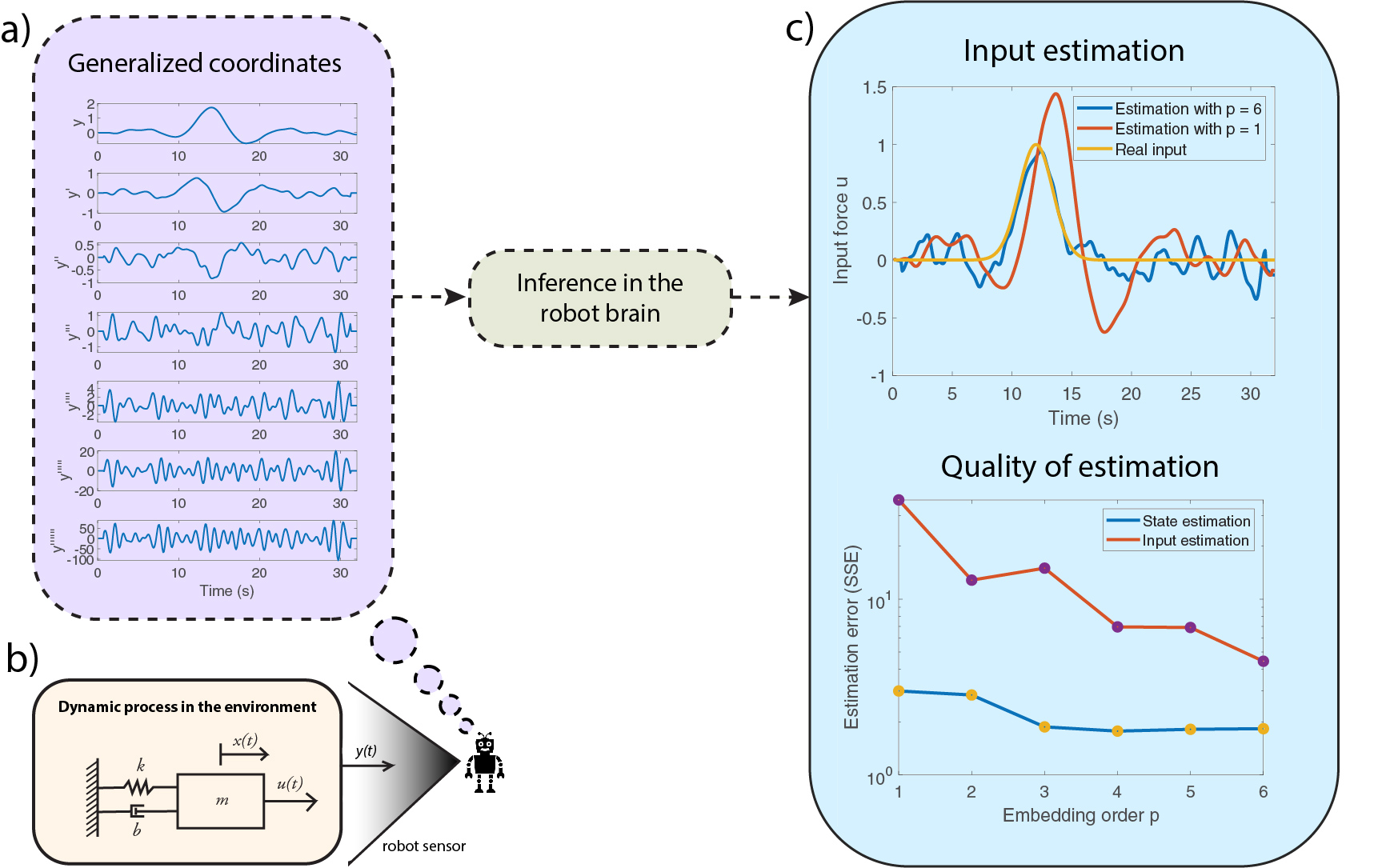}
\caption{An illustration of an attention mechanism in state and input estimation. The quality of the estimation improves as the embedding order (number of derivatives) of generalised coordinates are increased. However, the imprecise information in the higher order derivatives of the sensory input $\textbf{y}$ does not affect the final performance of the observer because of attentional selection, which selectively weighs the importance afforded to each derivative, in the free energy optimisation scheme.}
\label{fig:generalized_coordinates}
\end{figure}

Figure \ref{fig:generalized_coordinates} summarises a numerical example that shows how one can use precision modulation to focus on the less noisy derivatives (lower derivatives) of measurements, relative to imprecise higher derivatives. Thus, enabling the robot to use the most informative data for state and input estimation, while discarding imprecise input. Figure \ref{fig:generalized_coordinates}b depicts the mass-spring damper system used. The numerical results show that the quality of the estimation increases as the embedding ordering increases but the lack of information in the higher order derivatives of the sensory input do not affect the final performance due to the precision modulation. The higher order derivatives (Fig. \ref{fig:generalized_coordinates}a) are less precise than the lower derivatives, thereby reflecting the loss of information in higher derivatives. The state and input estimation was performed using the optimisation framework described in the previous section. The quality of estimation is shown in Fig. \ref{fig:generalized_coordinates}c, where the input estimation using six derivatives (blue curve) is closer to the real input (yellow curve) than when compared to the estimation using only one derivative (red curve). The quality of the estimation reports the sum of squared error (SSE) in the estimation of states and inputs with respect to the embedding order (number of signal derivatives considered). 

To obtain accurate state estimation by optimising the precision parameters, we recall that the precision weights the prediction errors. From Eq. (\ref{eqn:epsilon_precision}), the structural form of $\tilde{\Pi}$ is mainly dictated by the smoothness matrix $S$, which establishes the interdependence between the components of the variable expressed in generalised coordinates (e.g., the dependence between $\textbf{y}$, $\textbf{y}'$ and $\textbf{y}''$ in $\tilde{\textbf{y}}$). For instance, the $S$ matrix for a Gaussian kernel is as follows:
\begin{equation} \label{eqn:S_matrix}
    S = \begin{bmatrix}
    \frac{35}{16} & 0 & \frac{35}{8}s^2 & 0 & \frac{7}{4} s^4 & 0 & \frac{1}{6} s^6 \\
    0 & \frac{35}{4} s^2 & 0 & 7s^4 & 0 & s^6 & 0 \\
    \frac{35}{8}s^2 & 0 & \frac{77}{4}s^4 & 0 & \frac{19}{2} s^6 & 0 & s^8 \\
    0 & 7s^4 & 0 & 8s^6 & 0 & \frac{4}{3}s^8 & 0 \\
    \frac{7}{4} s^4 & 0 & \frac{19}{2} s^6 & 0 & \frac{17}{3}s^8 & 0 & \frac{2}{3}s^{10} \\
    0 & s^6 & 0 & \frac{4}{3} s^8 & 0 & \frac{4}{15}s^{10} & 0 \\
    \frac{1}{6} s^6 & 0 & s^8 & 0 & \frac{2}{3}s^{10} & 0 & \frac{4}{45}s^{12} \\
    \end{bmatrix},
\end{equation}
where $s$ is the kernel width of the Gaussian filter that is assumed to be responsible for serial correlations in measurement or state noise. Here, the order of generalised coordinates (number of derivatives under consideration) is taken as six ($S \in \mathbb{R}^{7 \times 7}$). For practical robotics applications, the measurement frequency is high, resulting in $0<s<1$. It can be observed that the diagonal elements of $S$ decreases because $s<1$, resulting in a higher attention (or weighting) on the prediction errors from the lower derivatives when compared to the higher derivatives. The higher the noise colour (i.e., $s$ increases), the higher the weight given to the higher state derivatives (last diagonal elements of $S$ increases). This reflects the fact that smooth fluctuations have more information content in their higher derivatives. Having established the potential importance of precision weighting in state estimation, we now turn to the estimation (i.e., learning) of precision in any given context.

\subsubsection{System identification} \label{sec:system_ID}
This section shows how to optimise system identification by means of precision learning~\citep{anil2021dynamic,10.1007/978-3-030-93736-2_49}. Specifically, we show how to fuse prior knowledge
about the dynamic model with the data to recover unknown parameters of the system through an attention mechanism.
This involves the learning of the 1) parameters and 2) noise precisions. Our model `turns' the attention to the least precise parameters and uses the data to update those parameters to increase their precision. Hence, allowing faster parameter learning. 

For the sake of clarity, we use again the mass-spring-damper system as the driving example (Sec. \ref{sec:prec_state_space}). We formalise system identification as evaluating the unknown parameters $k$, $m$ and $b$, given the input $\textbf{u}$, the output $\textbf{y}$, and the general form of the linear system in Eq. (\ref{eqn:spring_damper_LTI}).


\begin{figure}[!htb]
\centering
\includegraphics[scale = 0.23]{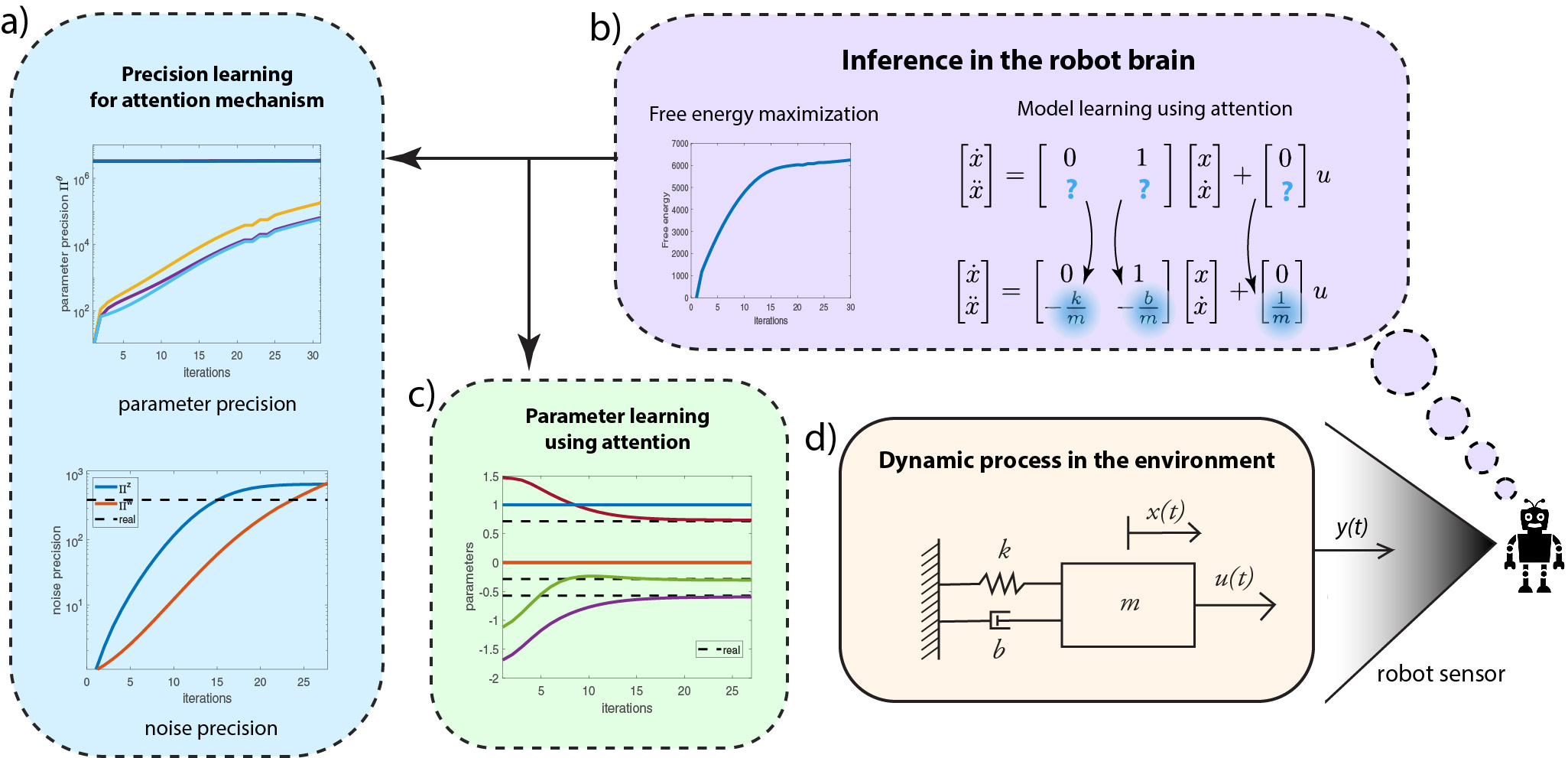}
\caption{The schematic of the robot's attention mechanism for learning the least precise parameters of a given generative model of a mass-spring-damper system. }
\label{fig:spring_damper_attention}
\end{figure}

Figure \ref{fig:spring_damper_attention} depicts the process of learning unknown parameters (dotted boxes denote the processes inside the robot brain). The robot measures its position $x(t)$ using its sensors (e.g., vision or range sensor). We assume that the robot has observed the behaviour of a mass-spring-damper system before or a model is provided by the expert designer. However, some of the parameters are unknown. The robot can reuse the prior learned model of the system to relearn the new system. This can be realised by setting a high prior precision on the known parameters and a low prior precision on the unknown parameters. By means of precision learning, the robot uses the sensory signals to learn the parameter precision $\Pi^\theta$, thereby improving the confidence in the parameter estimates $\theta$.  This directs the robot's attention towards the refinement of the parameters with least precision as they are the most uncertain. The requisite parameter learning proceeds by the gradient ascent of the free energy functional given in Eq. (\ref{eqn:F_action_opt}). The parameter precision learning proceeds by tracking the negative curvature of $\Bar{\mathcal{F}}$ as $\Pi^\theta = -\frac{\partial^2 \Bar{\mathcal{F}}}{\partial \theta^2} $ ~\citep{anil2021dynamic}.

The learning process -- by means of variational free energy optimisation (maximisation) -- is shown in Fig. \ref{fig:spring_damper_attention}b. The learning involves two parallel processes: precision learning (Fig. \ref{fig:spring_damper_attention}a), and parameter learning (Fig. \ref{fig:spring_damper_attention}c). Precision learning comprises of parameter precision learning (top graph) -- i.e., identifying the precision of an approximate posterior density for the parameters being estimated -- and noise precision learning (bottom graph). The high prior precision on the known system parameters (0 and 1), and low prior precision on the unknown system parameters ($-\frac{k}{m}, -\frac{b}{m}$ and $\frac{1}{m}$, highlighted in blue) directs attention towards learning the unknown parameters and their precision. Note that in Fig.~\ref{fig:spring_damper_attention}a, the precision on the three unknown parameters start from a low prior precision of $P^\theta=1$ and increase with each iteration, whereas the precision of known parameters (0 and 1) remains a constant ($3.3\times10^6$). The noise precisions are learned simultaneously, which starts from a low prior precision of $P^{\lambda^w}=P^{\lambda^z}=1$ and finally converges to the true noise precision (dotted black line). Both precisions are used to learn the three parameters of the system (Fig.~\ref{fig:spring_damper_attention}b), which starts from randomly selected values within the range [-2,2] and finally converges to the true parameter values of the system ($\theta_3 = -\frac{k}{m} = -0.5714$, $\theta_4 = -\frac{b}{m} =-0.2857$ and $\theta_6 = \frac{1}{m} =0.7143$), denoted by black dotted lines. From an attentional perspective, the lower plot in (Fig.~\ref{fig:spring_damper_attention}a) is particularly significant here. This is because the robot discovers the data are more informative than initially assumed, thereby leading to an increase in its estimate of the precision of the data-generating process. This means that the robot is not only using the data to optimise its beliefs about states and parameters (system identification), it is also using these data to optimise the way in which it assimilates these data.

In summary, precision-based attention, in the form of precision learning, helps the robot to accurately learn unknown parameters by fusing prior knowledge with new incoming data (sensory measurements), and attending to the least precise parameters. 

\subsubsection{Precision-modulated exploration and exploitation in system identification} \label{sec:exploration_expl}

Exploration and exploitation in the parameter space can be advantageous to robots during system identification. Precision-based attention---here the prior precision---allows a graceful balance between the two, mediated by the prior precision\footnote{Note that here we are using exploration and exploration not in terms of behaviour but for parameter learning. Exploration means adapting the parameter to a different (unexplored) value and exploitation means keeping that value}. A very high prior precision encourages exploitation and biases the robot towards believing its priors, while a low prior precision encourages exploration and makes the robot sensitive to new information. 

\begin{figure}[!htb] 
\centering
\includegraphics[scale = 0.33]{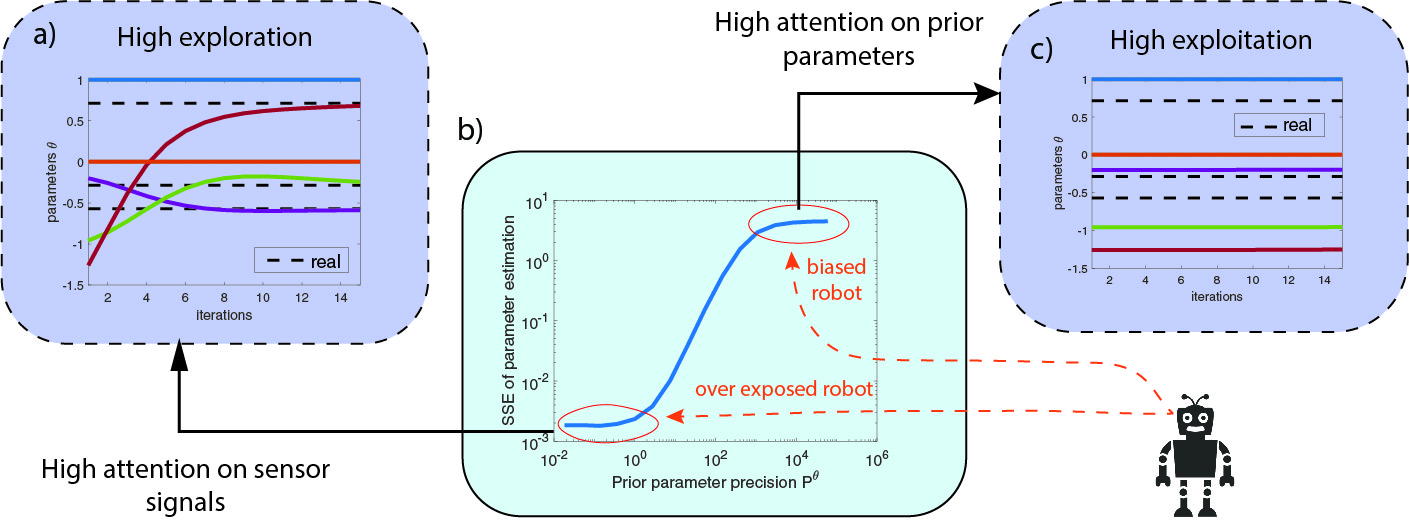}
\caption{Precision-based attention allows exploration and exploitation balanced model learning mediated by the prior precisions on the parameters $P^\theta$. The higher the $P^\theta$, the higher the attention on prior parameters $\eta^\theta$ and the lower the attention on the sensory signals while learning.}
\label{fig:exploration_exploitation}
\end{figure}

We use again the mass-spring-damper system example but with a different prior parameter precision $P^\theta$. The prior parameters are initialised at random and learned using optimisation. Figure \ref{fig:exploration_exploitation}b shows the increase in parameter estimation error (SSE) as the prior parameter precision $P^\theta$ increases until it finally saturates. The bottom left region (circled in red) indicates the region where the prior precision is
low, encouraging exploration with high attention on the sensory signals for learning the
model. This region over-exposes the robot to its sensory signals by neglecting the prior parameters. The top right region (circled in red) indicates the biased robot where the prior precision is high, encouraging the robot to exploit its prior beliefs by retaining high attention on prior parameters. This regime biases the robot
into being confident about its priors and disregarding new information from the sensory signals. Between those extreme regimes (blue curve) the prior precision balances the exploration-exploitation trade-off. Figure~\ref{fig:exploration_exploitation}a describes how increased attention to sensory signals helped the robot to recover from poor initial estimates of parameter values and converge towards the correct values (dotted black line). Conversely, in Fig. \ref{fig:exploration_exploitation}c, high attention on prior parameters did not help the robot to learn the correct parameter values. 

These results establish that prior precision modelling allows balanced exploration and exploitation of parameter space during system identification. Although the results show that an over-exposed robot provides better parameter learning, we show -- in the next section -- that this is not always be the case. 

\subsubsection{Noise estimation} \label{sec:noise_prec_learn}
\begin{figure}[!htb]
\centering
\includegraphics[scale = 0.3]{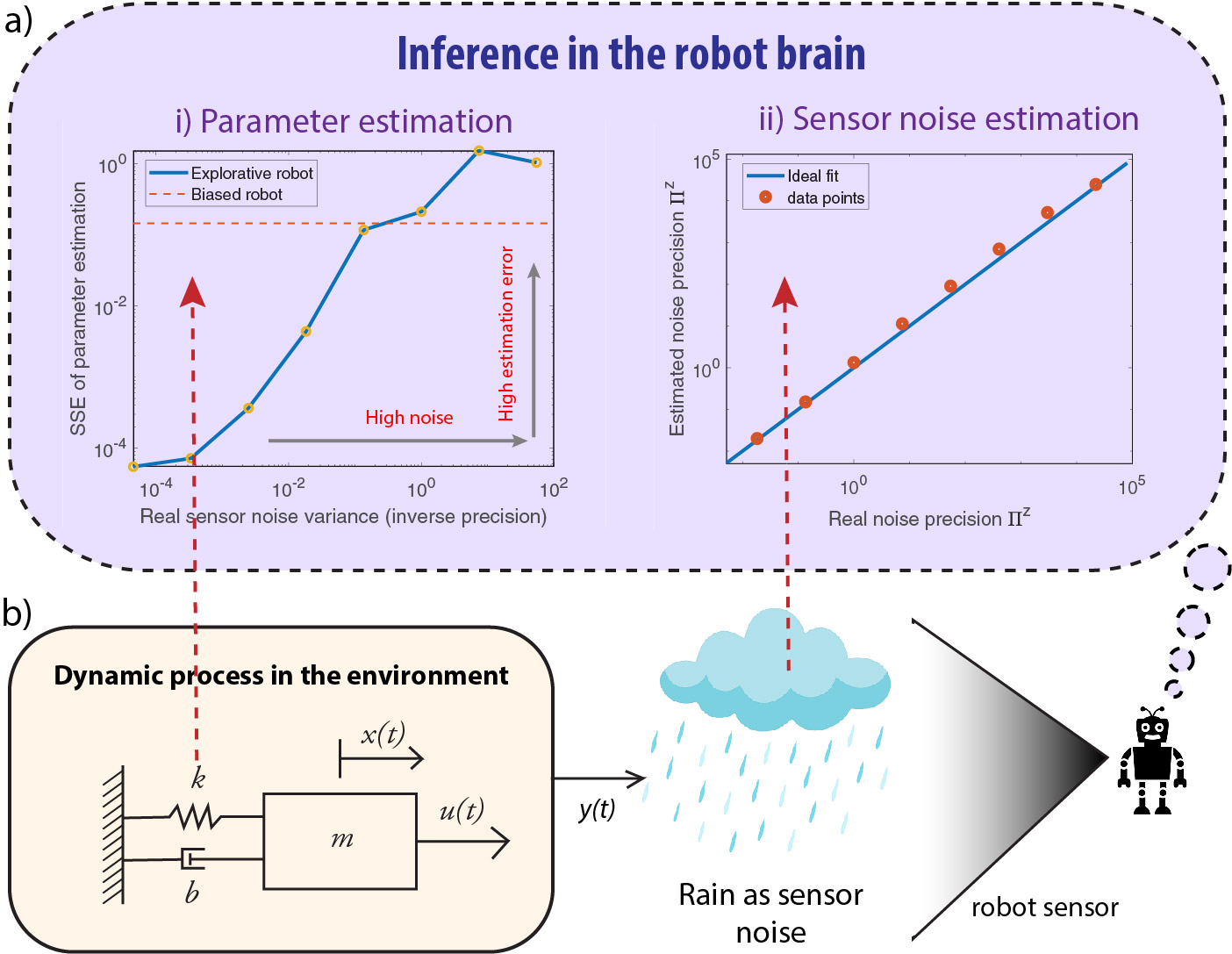}
\caption{Simulations demonstrating how a biased robot could be advantageous, especially while learning in a highly noisy environment. As the sensor noise increases, the quality of parameter estimation deteriorates to a point where an explorative robot generates higher parameter estimation errors than when compared to the biased robot that relies on its prior parameters. However, the sensor noise estimation is accurate even for high noise environments, demonstrating the success of the attention mechanism using the noise precision learning.}
\label{fig:good_biased_brain}
\end{figure}

In real-world applications, sensory measurements are often highly noisy and unpredictable. Furthermore, the robot does not have access to the noise levels. Thus, it needs to learn the noise precision ($\mathbf{\Pi^z}$) for accurate estimation and robust control. Precision-based attention enables this learning. In what follows, we show how one can estimate
 $\mathbf{\Pi^z}$ using noise precision learning and that biasing the robot to prior beliefs can be advantageous in highly noisy environments.

Consider again the mass-spring-damper system in Figure \ref{fig:good_biased_brain}b, where heavy rainfall/snow corrupts visual sensory signals. We evaluate the parameter estimation error under different noise conditions, using different levels of noise variances (inverse precision). For an over-exposed robot (only attending to sensory measurements), left plot of Fig. \ref{fig:good_biased_brain}a, the estimation error increases as the noise strength increases, to a point where the error surpasses the error from a prior-biased robot. This shows that a robot, confident in its prior model, assigns low attention to sensory signals and outperforms an over-exposed robot that assigns high attention to sensory signals, in a highly noisy environment. The right plot of Fig. \ref{fig:good_biased_brain}a shows the quality of noise precision learning for an over-exposed robot. It can be seen that all the data points in red lie close to the blue line, indicating that the estimated noise precision is close to the real noise precision. Therefore, the robot is capable of recovering the correct sensory noise levels even when the environment is extremely noisy, where accurate parameter estimation is difficult. 

These numerical results show that attention mechanism –- by means of noise precision learning –- allows the estimation of the noise levels in the environment and thereby protects against over-fitting or overconfident parameter estimation.

\noindent\textbf{Summary}. We have shown how precision-based attention—through precision modelling and learning—
yields to accurate robot state estimation, parameter identification and sensory noise estimation. In the next
section, we discuss how action is generated in this framework.

\subsection{Precision-modulated action}
\label{sec:rob:actions}

Selecting the optimal sequence of actions to fulfil a task is essential for robotics~\citep{lavalle2006planning}. One of the most prominent challenges is to ensure robust behaviour given the uncertainty emerging from a highly complex and dynamic real world, where the robots have to operate on. A proper attention system should provide action plans that resolve uncertainty and maximise information gain. For instance, it may minimise the information entropy, thereby encouraging repeated sensory measurements (observations) on high uncertainty sensory information. 

Salience, which in neuroscience is sometimes identified as Bayesian surprise (i.e., divergence between prior and posterior), describes which information is relevant to process. We go one step further by defining the saliency map as the epistemic value of a particular action~\citep{friston2015active}. 
Thus, the (expected) divergence now becomes the mutual information under a particular action or plan. This makes the saliency map more sophisticated because it is an explicit measure of the reduction in uncertainty or mutual information associated with a particular action (i.e., active sampling), and more pragmatic because it tells you where to sample data next, given current Bayesian beliefs.

\begin{figure}[!htb]
\centering
\captionsetup{justification=justified,margin=0cm}
\includegraphics[scale = 0.18]{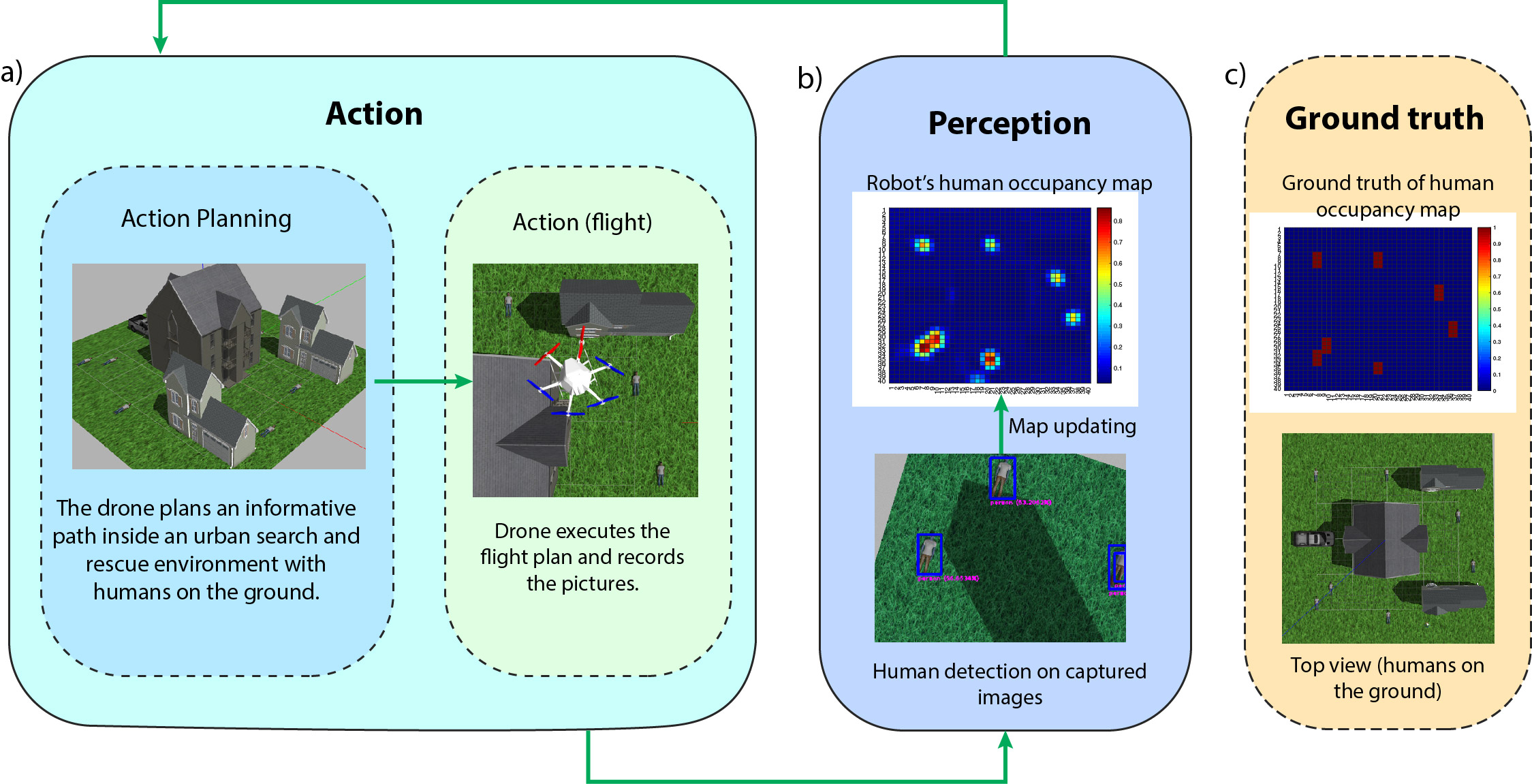}
\caption{IPP problem for localising human victims in an urban search and rescue scenario~\citep{meera2019obstacle}. A UAV, in a realistic simulation environment, plans a finite look-ahead path to minimise the uncertainty of its human occupancy map (e.g., modelled as a Gaussian process) of the world. The planned path is then executed, during which the UAV flies and captures images at a constant measurement frequency. After the data acquisition is complete, a human detection algorithm is executed to detect all the humans on the images. These detections are then fused into the UAV's human location map. The cycle is repeated until the uncertainty of the map is completely resolved (this usually implies enough area coverage and repeated measurements on uncertain locations). The ground truth of the human occupancy map and the UAV belief is shown in (c) and (b) respectively. The final map approaches the ground truth and all the seven humans on the ground are correctly detected.}
\label{fig:IPP_model}
\end{figure}

We first describe a precision representation usually used in information gathering problems and then how to directly generate action plans through precision optimisation. Afterwards, we discuss the realisation of the full-fledged model presented in the neuroscience section for active perception. We use the informative path planning (IPP) problem, described in Fig.~\ref{fig:IPP_model}, as an illustrative example to drive intuitions.

\subsubsection{Precision maps as saliency}
\label{sec:prec_info_maps}
One of the popular approaches in information gathering problems is to model the information map as a distribution (e.g., using Gaussian processes~\citep{hitz2017adaptive}). This is widely used in applications, such as a target search, coverage and navigation. The robot keeps track of an occupancy map and the associated uncertainty map (covariance matrix or inverse precision). While the occupancy map records the presence of the target on the map, the uncertainty map records the quality of those observations. The goal of the robot is to learn the distribution using some learning algorithm \citep{marchant2014bayesian}. A popular strategy is to plan the robot path such that it minimises the uncertainty of the map in future \citep{popovic2017multiresolution}. In Sec. \ref{sec:salience_uncertainty}, we will show how we can use the map precision to perform active perception, i.e., optimise the robot path for maximal information gain. Optimising the map precision drives the robot towards an exploratory behaviour.

\subsubsection{Precision optimisation for action planning}
\label{sec:rob:actions:optimization}
\begin{figure}[!hbtp]
\centering
\includegraphics[scale = 0.3]{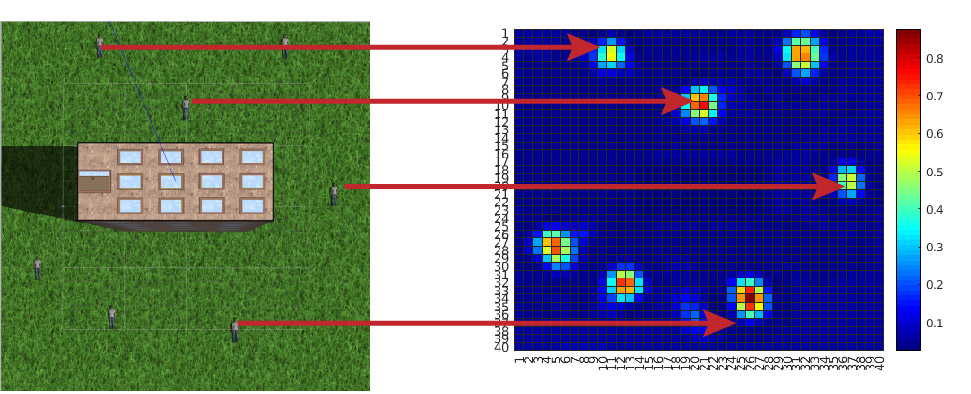}
\caption{Finding humans with unmanned air vehicles (UAVs): an informative path planning (IPP) approach~\citep{anil2018informative}. The simulation environment on the left consists of a tall building at the centre, surrounded by seven humans lying on the floor. The goal of the UAV is to compute the action sequence that allows maximum information gathering, i.e., the humans location uncertainty is minimised. On the right is the final occupancy map coloured with the probability of finding a human at that location. It can be observed that all humans on the simulation environment were correctly detected by the robot.}
\label{fig:human_detections}
\end{figure}\label{sec:salience_uncertainty}

\begin{figure}[!htb]
\centering
\includegraphics[scale = 0.4]{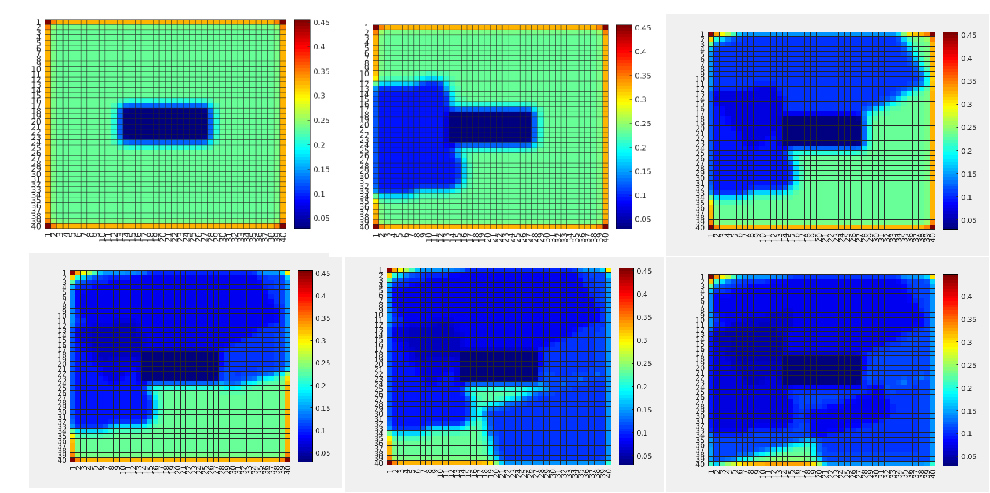}
\caption{Variance map of the probability distribution of people location (Fig. \ref{fig:human_detections}) -- inverse precision of human occupancy map. The plot sequence shows the  reduction of map uncertainty (inverse precision) after measurements~\citep{anil2018informative}. }
\label{fig:uncertainty_reduction}
\end{figure}

To introduce precision-based saliency we use an exemplary application of search and rescue. The goal is to find all humans using an unmanned air vehicle (UAV)~\citep{lanillos2013minimum,lanillos2014multi,rasouli2020attention,meera2019obstacle}. We use precision for two purposes: i) precision optimisation for action planning (plan flight path) and ii) precision learning for map refinement. In contrast to previous models of action selection within active inference in robotics~\citep{oliver2021empirical,lanillos2021-aif-challenges} here precision explicitly drives the agent behaviour. Figure \ref{fig:human_detections} describes the scenario in simulation. The seven human targets on the ground are correctly identified by the UAV. We can formalise the solution as the UAV actions (next flight path) that minimise the future uncertainties of the human occupancy map. In our precision-based attention scheme, this objective is equivalent to maximising the posterior precision of the map. Figure \ref{fig:uncertainty_reduction} shows the reduction in map uncertainty after subsequent assimilation of the measurements (camera images from the UAV, processed by a human detector).  The map (and precision) is learned using a recursive Kalman Filter by fusing the human detector outcome onto the map (and precision). The algorithm drives the UAV towards the least explored regions in the environment, defined by the precision map.

\begin{figure}[!htb]
\centering
\includegraphics[scale = 0.5]{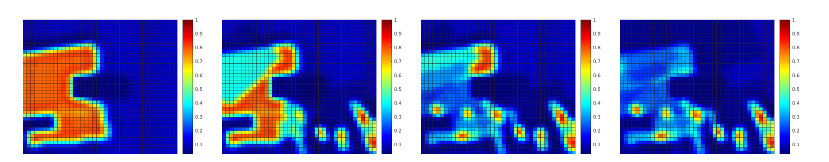}
\caption{The human occupancy map (probability to find humans at every location of the environment) at four time instances during the UAV flight showing ambiguity resolution. The ambiguity arising from imprecise sensor measurements (false positive) is resolved through repeated measurements at the same location. The plot sequence shows how the assimilation of the measurements updates the probability of the people being in each location of the map~\citep{meera2019obstacle}. }
\label{fig:sensor_attention}
\end{figure}

Furthermore, Fig. \ref{fig:sensor_attention} shows an example of uncertainty resolution under false positives. In this case, human targets are moved to the bottom half of the map. The first measurement provides a wrong human detection with high uncertainty. However, after repeated measurements at the same location in the map the algorithm was capable of resolving this ambiguity, to finally learn the correct ground truth map. Hence, the sought behaviour is to take actions that encourage repeated measurements at uncertain locations for reducing uncertainty.  

Although the IPP example illustrates how to generate control actions through precision optimisation, the
task, by construction, is constrained to explicitly reduce uncertainty. This is similar to the description of visual search described in~\citep{friston2012perceptions}, where the location was chosen to maximise information gain. Information gain (i.e., the Bayesian surprise expected following an action) is a key part of the expected free energy functional that underwrite action selection in active inference. In brief, expected free energy can be decomposed into two parts: the first corresponds to the information gain above (a.k.a., epistemic value or affordance), and the second corresponds to the expected log evidence or marginal likelihood of sensory samples (a.k.a., pragmatic value). When this likelihood is read as a prior preference, it contextualises the imperative to reduce uncertainty by including a goal-directed imperative. For example, in the search paradigm above, we could have formulated the problem in terms of reducing uncertainty about whether each location was occupied by a human or not. We could have then equipped the agent with prior preferences for observing humans. 

In principle, this would have produced searching behaviour until uncertainty had been resolved about the scene; after which, the robot would seek out humans; simply because, these are its preferred outcomes. In thinking about how this kind of neuroscience inspired or biomimetic approach could be implemented in robotics, one has to consider carefully, the precision afforded sensory inputs (i.e., the likelihood of sensory data, given its latent causes) – and how this changes during robotic flight and periods of data gathering. This brings us back to the precision modulation and the temporal scheduling of searching and securing data. In the final section, we conclude with a brief discussion of how this might be implemented in future applications.

\subsubsection{Precision-based active perception} \label{sec:act_per_coupling}

\begin{figure}[!hbtb]
\centering
\includegraphics[scale = 0.23]{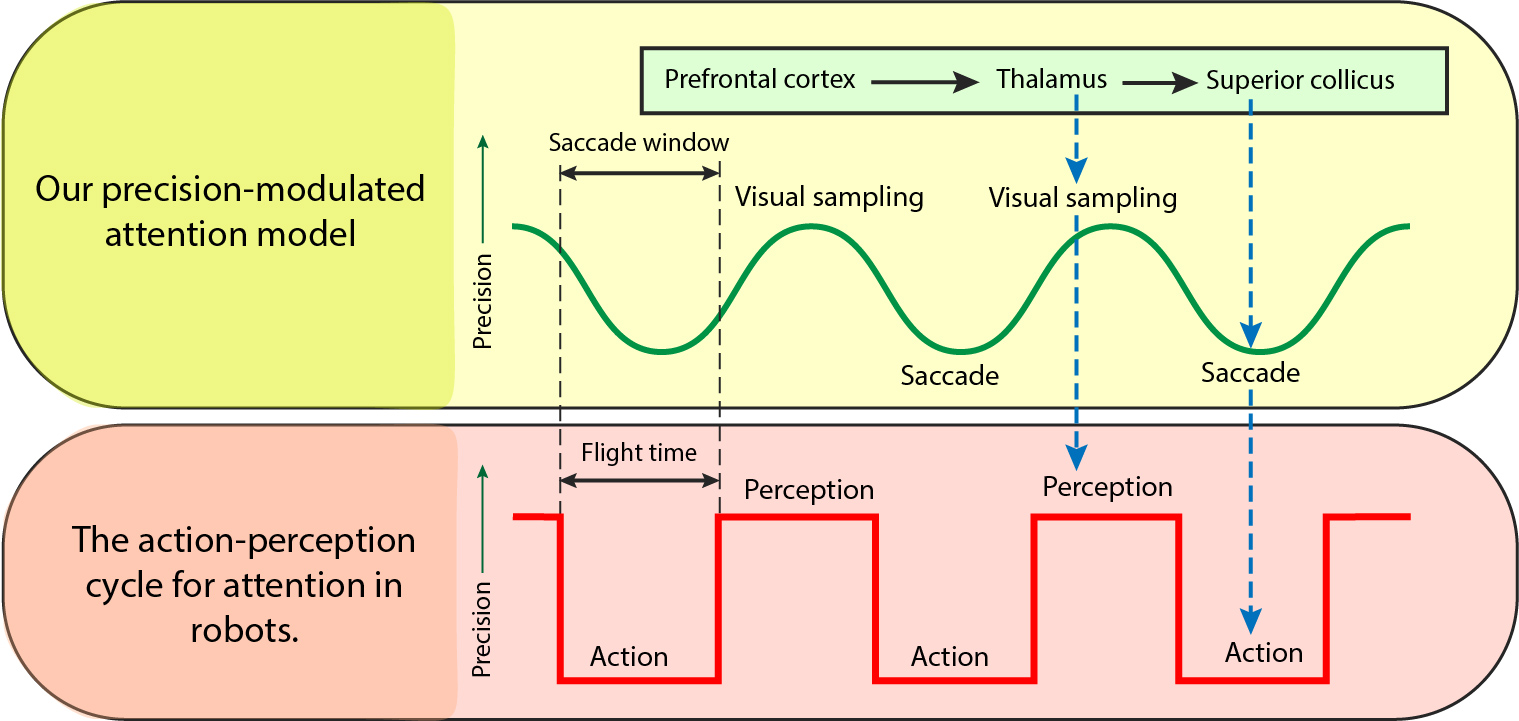}
\caption{Precision-modulated attention model adapted to the action-perception loop in robotics. Each cycle consists of two steps: 1) action (planning and execution of a finite-time look ahead of the robot path for data collection) and 2) perception (learning using the collected data). This scheduling, using a finite time look-ahead plan, is quite common in real applications and of particular importance when processing is computationally expensive, e.g., slow rate of classification, non-scalable data fusion algorithms, Exponential planners, etc. However, the benefits of incorporating 'optimal' scheduled loop driven by precision should be further studied.
}
\label{fig:salience_pre_model}
\end{figure}
In this section, we discuss the realisation of a biomimetic brain-inspired model in relation to existing solutions in robotics in the context of path-planning. Figure \ref{fig:salience_pre_model} compares our proposed precision-modulated attention model---from Fig. \ref{fig:summary_fig}---with the action-perception loop widely used in robotics. By analogy with eye saccades to the next visual sample, the UAV flies (action) over the environment to assimilate sensory data for an informed scene construction (perception). Once the flight time of the UAV is exhausted (similar to saccade window of the eye), the action is complete, after which the map is updated, and the next flight path is planned. 

In standard applications of active inference, the information gain is supplemented with expected log preferences to provide a complete expected free energy functional~\citep{sajid2021efe}. This accommodates the two kinds of uncertainty that actions and choices typically reduce. The first kind of uncertainty is inherent in unknowns in the environment. This is the information gain we have focused on above. The second kind of uncertainty corresponds to expected surprise, where surprise rests upon a priori expected or preferred outcomes. As noted above, equipping robots with both epistemic and pragmatic aspects to their action selection or planning could produce realistic and useful behaviour that automatically resolves the exploration-exploitation dilemma. This follows because the expected free energy contains the optical mixture of epistemic (information-seeking) and pragmatic (i.e., preference seeking) components. Usually, after a period of exploration, the preference seeking components predominate because uncertainty has been resolved. Although expected free energy provides a fairly universal objective function for sentient behaviour, it does not specify how to deploy behaviour and sensory processing optimally. This brings us to the precision modulation model, inspired by neuroscientific considerations of attention and salience.

Hence, there is a key difference between biological and robotic implementations of the search behaviour, which is the use of continuous oscillatory precision to modulate visual sampling and movement cycles, as opposed to arbitrary discrete action and perception steps currently used in robotics. 
Importantly, our salience formulation speaks to selecting future data that reduces this uncertainty. For instance, we have shown---in the information gathering IPP example described in the previous subsection---that by optimising precision we also optimise behaviour.

We argue the potential need and the advantages of realising precision based temporal scheduling, as described in our brain-inspired model, for two practically relevant test cases: ($i$) learning dynamic models and ($ii$) information seeking applications. 

In Section \ref{sec:exploration_expl}, we have shown how the exploration-exploitation trade-off can be mediated by the prior parameter precision during learning. However, the accuracy-precision curve (Fig. \ref{fig:exploration_exploitation}b) is often practically unavailable due to unknown true parameters values, challenging the modelling of prior precision. An alternative would be to use a precision based temporal scheduling mechanism to alternate between exploration and exploitation by means of a varying $P^\theta$ (similar to Fig. \ref{fig:salience_pre_model}) during learning, such that system identification is neither biased nor over exposed to sensory measurements. In Fig. \ref{fig:good_biased_brain}a, we showed how noise levels influence estimation accuracy, and how biasing the robot by modelling $P^\theta$ can be beneficial for highly noisy environments. A precision based temporal scheduling mechanism by means of a varying $P^\theta$ could provide a balanced solution between a biased robot (that exploits its model) and an exploratory one. 

Furthermore, temporal scheduling, in the same way that eye saccades are generated, can be adapted for information gathering applications, such as target search, simultaneous localization and mapping, environment monitoring, etc. For instance, introducing precision-modulation scheduling for solving the IPP, and scheduling perception (map learning) and action (UAV flight). Precision modulation will switch between action and perception: when the precision is high, perception occurs (c.f., visual sampling), and when the precision is low, action occurs (c.f., eye movements). This switch, which is often implemented in the robotics literature using a budget for flight time, will be now dictated by precision dynamics.

In short, we have sketched the basis for a future realisation of precision-based active perception, where the robot computes the actions to minimise the expected uncertainty. While most attentional mechanisms in robotics are limited to providing a `saliency' map highlighting the most relevant features, our attention mechanism proposes a general scheduling mechanism with action in the loop with perception, both driven by precision.

\section{Concluding remarks}
\label{sec:conclusion}
We have considered attention and salience as two distinct processes that rest upon oscillatory precision control processes. Accordingly, they require particular temporal considerations: attention to
reliably estimate latent states from current sensory data and salience for uncertainty reduction regarding future data samples.
This formulation addresses visual search from a first principles (Bayesian) account of how these mechanisms might manifest -– and the circular causality that undergirds them via a rhythmic theta-coupling.
Crucially, we have revisited the definition of salience from the visual neurosciences; where it is read as Bayesian surprise (i.e., the Kullback Leibler divergence between prior and posterior beliefs). We took this one step further and defined salience as the expected Bayesian surprise (i.e., epistemic value) of a particular action (e.g., sampling this set of data)~\citep{friston2017active1,sajid2021efe}. Formulating salience as the expected divergence renders it 
the mutual information under a particular action (or action trajectory)~\citep{friston2021sophisticated}. For brevity, our narrative was centred around visual attention and its realisation via eye movements. However, this model does not strictly need to be limited to visual information processing, because it addresses sensorimotor and auditory processing in general. This means it explains how action and perception can be coupled in other sensory modalities. For instance, \citep{tomassini2017theta} showed that visual information is coupled with finger movements at a neural theta rhythm.

The point of contact with the robotics use of salience emerges because the co-variation between a particular parameterisation and the inputs is a measure of the mutual information between the data and its estimated causes. In this sense, both definitions of salience reflect the mutual information -- or information about a particular representation of a (latent) cause -- afforded by an observation or consequence. However, our formulation is more sophisticated. Briefly, because it is an explicit measure of the reduction in uncertainty (i.e., mutual information) associated with a particular action (i.e., active sampling) and specifies where to sample data next, given current Bayesian beliefs. These processes (attention and salience) are a consequence of precision of beliefs over distinct model parameters. Explicitly, attention contends with precision over the causes of (current) outcomes and salience contends with beliefs about the data that has to be acquired and precision over beliefs about actions that dictate it. Since both processes can be linked via precision manipulation, the crucial thing is the precision that differentiates whether the agent acquires new information (under high precision) or resolves uncertainty by moving (low precision). 

The focus of this work has been to illustrate the importance of optimising precision at various places in generative models used for data assimilation, system identification and active sensing. A key point – implicit in these demonstrations – rests upon the mean field approximation used in all applications. Crucially, this means that getting the precision right matters, because updating posterior estimates of states, parameters and precisions all depend upon each other. This may be particularly prescient for making the most sense of samples that maximises information gain. In other words, although attention and salience are separable optimisation processes, they depend upon each other during active sensing. This was the focus of our final numerical studies of action planning.

To face-validate our formulation, we evaluated precision-modulated attentional processes in the robotic domain. We presented numerical examples to show how precision manipulation underwrites accurate state and noise estimation (e.g., selecting relevant information), as well as allowing system identification (e.g., learning unknown parameters of the dynamics). We also showed how one can use precision-based optimisation to solve interesting problems; like the informative path planning in search and rescue scenarios. Thus, in contrast to previous uses of attention in robotics, we placed attention and saliency as integral processes for efficient gathering and processing of sensory information. Accordingly, `paying attention' is not only about filtering the current flow of information from the sensors but performing those actions that minimise expected uncertainty. Still, the full potential of our proposal has yet to be realised, as the precision-based attention should be able to account for prior preferences beyond the IPP problem (e.g., localising people using UAVs). Finally, we briefly considered the realisation of temporal scheduling for information gathering tasks, opening up interesting lines of research to provide robots with biologically plausible attention.

\section*{Author Contributions}
AAM and FN are responsible for the novel account and its translation to robotics. All authors contributed to conception and design of the work. AAM, FN, PL and NS wrote the manuscript. All authors contributed to manuscript revision, read, and approved the submitted version.

\section*{Funding}
P.L. is partially supported by Spikeference project, Human Brain Project Specific Grant Agreement 3 (ID: 945539). N.S. is funded by the Medical Research Council (MR/S502522/1) and 2021–2022 Microsoft PhD Fellowship. K.F. is supported by funding for the Wellcome Centre for Human Neuroimaging (Ref: 205103/Z/16/Z) and a Canada‐UK Artificial Intelligence Initiative (Ref: ES/T01279X/1).

\bibliographystyle{unsrtnat}
\bibliography{main}  






\end{document}